\documentclass[conference]{IEEEtran}
\IEEEoverridecommandlockouts
\usepackage{cite}
\usepackage{amsmath,amssymb,amsfonts}
\usepackage{graphicx}
\usepackage{textcomp}
\usepackage{booktabs}
\usepackage{tabularx,booktabs}
\usepackage{lipsum}
\usepackage{makecell}
\usepackage{algorithm}
\usepackage[utf8]{inputenc}
\usepackage{amsmath}
\usepackage{verbatim}
\usepackage[draft,colorinlistoftodos,color=orange!30]{todonotes}
\usepackage{color}
\newcommand{\draftnote}[3]{ 
	\todo[author=#2,color=#1!30,size=\footnotesize]{\textsf{#3}}	}
\newcommand{\qli}[1]{\draftnote{red}{QLi:}{#1}}
\renewcommand\IEEEkeywordsname{Keywords}

\def\BibTeX{{\rm B\kern-.05em{\sc i\kern-.025em b}\kern-.08em
    T\kern-.1667em\lower.7ex\hbox{E}\kern-.125emX}}
\begin{document}

\title{Causality Learning: A New Perspective for Interpretable Machine Learning}

\author{\IEEEauthorblockN{Guandong Xu,
Tri Dung Duong, Qian Li, 
Shaowu Liu and Xianzhi Wang}
\IEEEauthorblockA{School of Computer Science,
University of Technology Sydney, Australia\\
\{Guandong.Xu,Qian.Li,Shaowu.Liu,Xianzhi.Wang\}@uts.edu.au,
TriDung.Duong@student.uts.edu.au}}

\maketitle

\thispagestyle{plain}
\pagestyle{plain}

\begin{abstract}
Recent years have witnessed the rapid growth of machine learning in a wide range of fields such as image recognition, text classification, credit scoring prediction, recommendation system, etc. In spite of their great performance in different sectors, researchers still concern about the mechanism under any machine learning (ML) techniques that are inherently black-box and becoming more complex to achieve higher accuracy.
Therefore, interpreting machine learning model is currently a mainstream topic in the research community. However, the traditional interpretable machine learning focuses on the association instead of the causality. This paper provides an overview of causal analysis with the fundamental background and key concepts, and then summarizes most recent causal approaches for interpretable machine learning. The evaluation techniques for assessing method quality, and open problems in causal interpretability are also discussed in this paper.   
\end{abstract}


\section{Introduction}
In the past decades, machine learning has achieved the impressive performance in diverse tasks, and is increasingly applied in science, society and business. However, most of state-of-the-art models remained incomprehensible for both researchers, users and engineers, causing difficulties when deploying in real world. Specifically, there are several high-stake decision-making domains such as self-driving cars, crime prediction or personalized medicine in which the lack of transparency in machine learning prevents themselves from being adopted. Take for instance, in the healthcare sector where each decision can affect the people's survival, physicians are frequently concerned about the safety and trust of any deployed models. They do not likely trust the model's prediction if they can not understanding the rationales behind it. Consequently, interpretability in machine learning plays a significantly important role in generating trust-worthy models. This furthermore allows researchers, data scientists and engineers to ensure the models following the human understanding, ethnic codes, fairness and security. We as human have an insatiable curious nature; thus, our goal is not only to understand models' mechanism but also to generate and extract new knowledge of the world.

\begin{figure*}[!htb]
  \includegraphics[width=\linewidth]{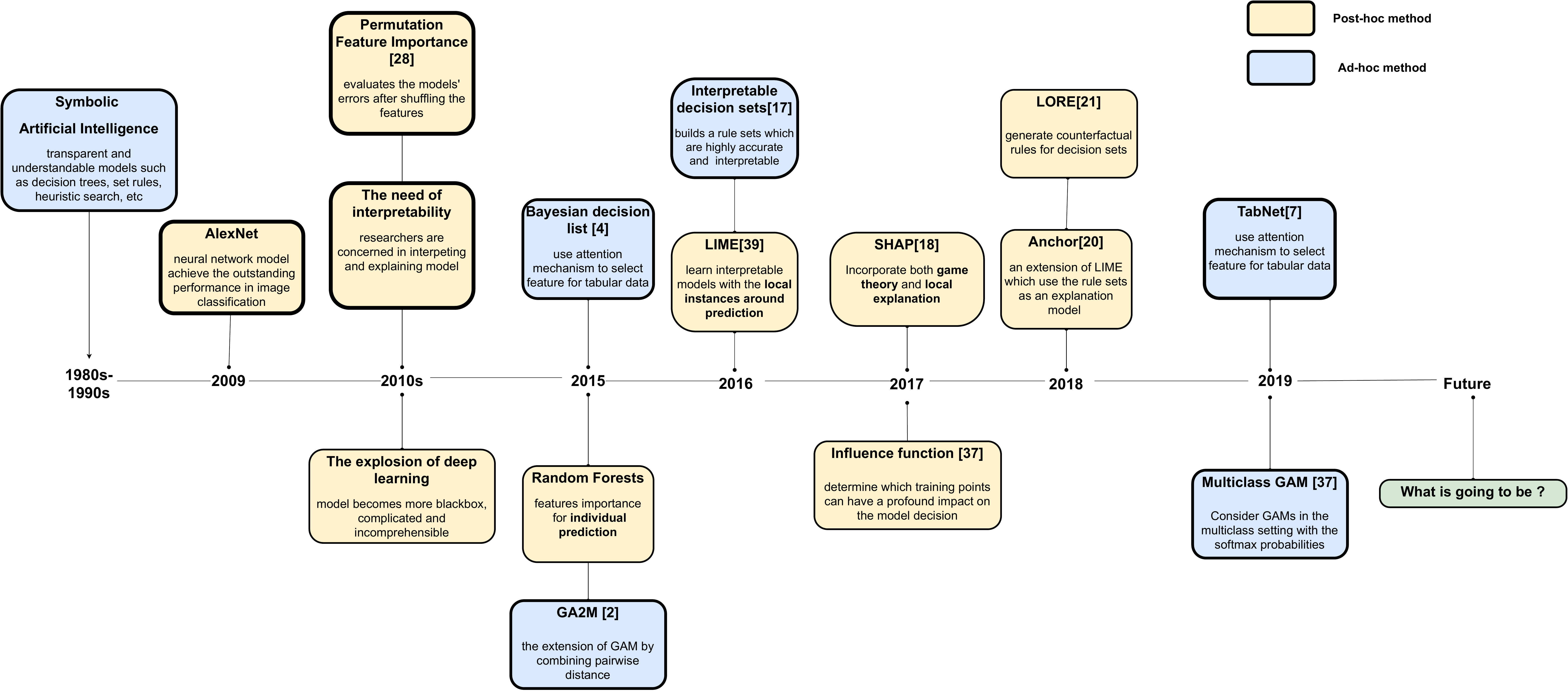}
  \caption{The evolution of interpretable machine learning}
  \label{fig:history}
\end{figure*}

In view of the time of explanation generation shown in Figure~\ref{fig:history}, interpretable machine learning can be divided into two branches: ad-hoc and post-hoc methods. The evolutionary history of noticeable traditional interpretable machine learning techniques is briefly described in the Figure \ref{fig:history}. The ad-hoc type focuses on building the model architecture, algorithms or mechanisms that are self-explainable and transparent.
Intrinsically interpretable models are the central research in the early years of artificial intelligence with the dominance of symbolism methods, followed by more advanced approaches such as decision sets \cite{lakkaraju2016interpretable}, generalized linear regression, generalized additive model \cite{zhang2019axiomatic, caruana_intelligible_2015,zhang_axiomatic_2019}, Bayesian probabilistic model \cite{darwen2019bayesian,letham2015interpretable}, rule-based model \cite{wang_multi-value_2017, letham_interpretable_2015}, attention mechanism \cite{arik_tabnet_2019}, fuzzy inference systems \cite{jang1993functional, guillaume2001designing, wang2002self}, TabNet \cite{arik_tabnet_2019}, etc. 
With the rapid growth of deep learning in recent decades, machine learning model is gradually evolved into complicated and incomprehensible form, which leads to the increasing attention on post-hoc interpretations.
Several prominent approaches in this category include Local surrogate models (LIME \cite{ribeiro_2016}, SHAP \cite{lundberg2017unified}, LORE \cite{guidotti2018local}, etc), influence functions \cite{koh2017understanding} and feature importance estimation \cite{schwab_granger-causal_2019, schwab_cxplain_2019} have been introduced. 

However, traditional interpretable machine learning focuses on the association instead of the causality.
With the emergence of causal inference, an increasing number of causality-oriented methods have been proposed in interpretable machine learning. In comparison with traditional methods, causal approaches can be utilized to identify causes and effects of models architecture or conduct the reasoning over its decisions and behaviors. 
This article examines the overview of interpretable machine learning, presents the causal analysis in machine learning interpretability and finally discusses the future research directions. More specifically, we first present the background of causal analysis with key concepts, models and evaluation metrics. We then provide an overview of state-of-the-art works on causal interpretability. We also illustrate the potential evaluation metrics used in interpretable machine learning. 

\section{Causality Analysis}

Causality analysis can exploit the causality mechanisms underlying the data-generating process, which is more advanced than the predictive or descriptive capability in machine learning techniques.
Causal inference and causal discovery are two main research topics for causality analysis. The goal of causal inference is to estimate the causal effect of treatment (i.e., a decision made or action taken) on the outcome (i.e., the result of treatment). Causal discovery examines whether a set of causal relationships exists among the variables.
This paper would primarily focus on causal effect, which is more correlated to machine learning interpretability.

\subsection{Causal Inference}
Causal inference has been widely applied in econometric, social science and medicine fields for evaluating the policy’s effect or the drugs’ side effect. Effect estimation is tied to the outcome caused by the treatment applied to an instance. 
An instance is the atomic research object, which can be a physical object or an individual person. Treatment and outcome are terms that denote a decision made or action taken and its result, respectively.
We first introduce the essential concepts for learning treatment effect followed by the causal models.

\begin{itemize}
    \item Covariates $X$ refers to the background variables or features of the instance.
    \item Treatment $T$ refers to the action (manipulation or intervention) that applies to a instance.
    \item Outcome $Y$ is the result of the treatment applied on a instance.
    \item Confounder $Z$ is a variable which causally affects both treatment and outcome. 
\end{itemize}

\begin{figure}[!htb]
  \includegraphics[width=\linewidth,scale=0.05]{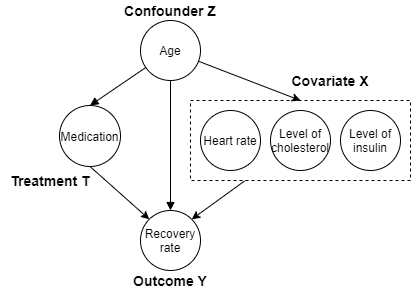}
  \caption{The causal graph for recovery rate problem}
  \label{fig:causal_graph}
\end{figure}

To better understand causal inference, we give the following example combined with the notations defined above. 
To prove the efficiency of the medication on the disease, the scientist needs to assess its positive effect into the patients' recovery rate. Figure~\ref{fig:causal_graph} depicts the corresponding causal relationships among the essential variables. The treatment $T$ is whether the drugs are applied or not, and the observed features $X$ are the patients' condition such as the level of insulin and cholesterol, heart rate, etc. Outcome $Y$ is the recovery rate and age is the confounder $Z$. This is simply because age firstly determined the need of applying medication into patients, since the young people may not necessarily take the medicine. Age also affects to the recovery rate: the youth has a higher probability to recover than the elderly. 

\subsection{Causal Models}
We now introduce the two most important formal frameworks used for causal inference, namely the structural causal models and the potential outcome framework. 


\textbf{Structural causal model}\cite{article_causal} consists of two main components: the causal graph and structural equations. Causal graph is the probabilistic graphical model which is used to represent the assumption about prior knowledge and data generating process. A causal graph is defined as $\mathcal{G} = \langle \mathcal{V}, \mathcal{E} \rangle$ where $\mathcal{V}$ is the set of nodes and $\mathcal{E}$ is the set of edges. 
Structural equation is a set of equations Eq. \eqref{eqn:structural_equation} which are used to represent the causal effect illustrated by the edge in the causal graph.  
    \begin{align}
    \label{eqn:structural_equation}
    \begin{split}
      X = f_X(E_X) ,
    \\
     T = f_T(X, E_T) . 
     \\
     Y = f_Y(X, D, E_Y)
    \end{split}
    \end{align}
where $E_X, E_T, E_Y$ are exogenous variables, which are independent from other models' variable, and are determined outside the model.




\textbf{Potential outcome framework} is proposed by Neyman and Rubin~\cite{rubin1974estimating}. 
Considering binary treatments for a set of units, there are two possible outcomes for each unit.
The unit will be assigned to the control treatment if $T=0$, 
or to the treated treatment if $T=1$. 
As a result, we denote two potential outcomes $Y_0$ and $Y_1$ as the results caused by $T=0$ and $T=1$, respectively.
Importantly, only one potential outcome is observed corresponds to the assigned treatment $T$, 
and we call this as the observed (factual) outcome $Y$. The unobserved potential outcome refers to the counterfactual outcome. 
Given the treatment $T_i$, 
the relationship between the observed outcome $Y$ and two potential outcomes are
\begin{equation}
\begin{split}
Y_i= T_iY_{1}+ (1-T_i)Y_{0}
\end{split}
\label{eq:y_F}
\end{equation}

\subsection{Treatment Effect Metric}
With the key concepts and causal models, the treatment effect can be measured at the population, treated group, subgroup, and individual level.
For simplicity, we discuss the treatment effect under the binary treatment, and it can be easily extended to multiple treatments by considering multiple potential outcomes.

The individual treatment effect (ITE) is defined as the change of $Y_0$ and $Y_1$,
while keeping the covariates $X$ unchanged (i.e., condition on those covariates). For an instance $i$ with covariates $X_i$, its corresponding ITE is 
\begin{equation}
ITE (\boldsymbol{X}_i)=E[Y_1|X_i]-E[Y_0|X_i]
\label{eq:ite_g}
\end{equation}
As only one potential outcome is observed, it is nearly impossible to estimate the effect at the individual level. A more feasible way is to measure treatment effect at the average level.

The average treatment effect (ATE) measures the treatment effect at the whole population level as 
        \begin{equation} 
        \label{eq1}
        \begin{split}
        ATE & = E[Y_1 - Y_0]
        \end{split}
        \end{equation}

The average treatment effect (ATT) is for the group of instances with the treatment equal to 1, i.e., the treated group.
        \begin{equation} 
        \label{eq1}
        \begin{split}
        ATT & = E[Y_1 - Y_0 | T = 1]
        \end{split}
        \end{equation}

Conditional average treatment effect (CATE) known as heterogeneous treatment effect is defined on the subgroup with the particular covariate $X=x$.
        \begin{equation} 
        \label{eq1}
        \begin{split}
        CATE & = E[Y_1 - Y_0 | X = x]
        \end{split}
        \end{equation}

\subsection{Tools for Causal Analysis}
Several libraries or tools are available for causal inference. Examples including \emph{Double Machine Learning} \cite{chernozhukov2016double}, \emph{Meta-learners} \cite{RN165}, \emph{Orthogonal Learning} \cite{oprescu2018orthogonal, foster2019orthogonal} have been supported by EconML, CausalML, DoWhy and CausalNex, whereas causal discovery methods including graph inference and pairwise inference are provided in Causal Discovery Toolbox. Meanwhile, TIGRAMITE is a novel framework for causal discovery in time series. 
We summarize the existing toolboxes in Table~\ref{fig:toolbox}.

\section{Interpretable machine learning with causality}


Pearl~\cite{10.5555/3238230} argues that causal reasoning is indispensable for machine learning to reach the human-level artificial intelligence, since it is the basic mechanism of human to be aware of the world. As a result, causal methodology is gradually becoming a vitally important component in explainable and interpretable machine learning. However, most of current interpretability techniques pay attention to solving the correlation statistic rather than the causation. Therefore, the causal approaches should be emphasized to achieve a higher degree of interpretability.


\subsection{Model-Agnostic Causality for Deep Neural Neworks}
The traditional way to analyze Deep Neural Network is to build several models with different architectures and make a comparison between their performances. The problem is that re-training DNNs is computationally expensive, and infeasible when it comes to the complicated architecture. Inspired by causal model, several methods have been proposed to interpret neural network model.

Chattopadhyay et al. \cite{chattopadhyay2019neural} define $\text{ACE}^{y}_{do(x_i = \alpha)}$ as the causal attribution of neuron $x_i$ to the output neuron $y_i$, and $\mathbb{E}[y|do(x_i = \alpha)]$ as the interventional expectation Eq. \eqref{causal_att}. The polynomial function is selected to estimate this value.

\begin{equation}
    \label{causal_att}
    \mathbb{E}[y|do(x_i = \alpha)] = \int yp(y|do(x_i = \alpha)) dy
\end{equation}


Narendra et al. \cite{narendra2018explaining} propose to construct a modified structural causal model as an abstraction of a DNN to make an reasoning over its elements. Thereafter, they rank each component based on their contribution to the final prediction for evaluation. 

Based on TCAV \cite{Kim2017InterpretabilityBF} which generates a high-level concept-based explanation such as gender, race, background, others, the study in \cite{goyal_explaining_2020} evaluates the \emph{causal concept effect} on a neural network prediction. They overcome the problem of do-operator by using Variational AutoEncoder (VAE).

Regarding Generative Adversarial Networks (GANs) interpretability, Bau et al.~\cite{bau2018gan} proposes an approach for visualization and understanding at unit-, object-, and scene- level by estimating the causal effect of the models' interpretable components. There are two main steps in their approach: dissection and intervention. In the dissection step, the classes with the explicit representation are firstly identified. Thereafter, they make an intervention by forcing the units to be appeared and disappeared, and calculate its causal effect. Meanwhile, the authors \cite{besserve2018counterfactuals} propose a causal framework to explore the intervention effect for proving that the components in images generated by GAN can be modified independently. 

In terms of reinforcement learning, action influence model \cite{madumal2019explainable} is introduced for explaining the behavior of RL agents. They construct a modified structural causal model, learn the causal equation as the regression model during training the agent, and finally generate the contrastive explanation to answer the counterfactual question "Why does the agent choose action A instead of action B?".

\subsection{Post-hoc Interpretability}

Model-Agnostic explanations are particular challenging when the models' parameters have more complex relationships. To further aid the intepretability, the practitioners propose a variety of post-hoc interpretability methods to exploit what a trained model has learned, without changing the underlying model. Most widely useful post-hoc interpretation methods fall into two main categories: causal feature learning and counterfactual explanations, respectively. 

\subsubsection{Casual Feature Learning}

Recent work on feature learning derives the subset of features that have causal contributions to the models' prediction. Early causal feature learning is to find the Markov Blanket (MB) containing a set of features which makes the target (T) independent from other features given MB(T). In the study \cite{pmlr-v3-cawley09a}, the authors firstly use the HITON algorithm \cite{aliferis2003hiton} to derive the Markov Blanket, and thereafter deploy Max-Min Hill-Climbing (MMHC) algorithm to identify the causes and effects of the target variable. Given the number of transfer learning tasks $D$, Peters et al. \cite{2015arXiv150101332P} assume that there exists a subset of features $X_{S^*}$ such that the conditional distribution $Y_k | X_{S*}$ is the same for different tasks $k$, and other settings Eq. \eqref{eqn:conditional}. They propose an algorithm called subset search which samples the subset features, and then adopt the Levene test to assess the assumption.

\begin{equation}
\label{eqn:conditional}
    Y_k | X_{S*}^k \approx Y_k' | X_{S*}^{k'} \quad \forall k, k' \in {1,...,D}
\end{equation}

CXPlain \cite{schwab_cxplain_2019} is the causal framework that can explain more complex machine learning models by estimating the feature importance. Granger-causal objective is introduced to quantify how much the exclusion of a single feature reduces model performance. Particularly, CXPlain trains a separate explanation model to any predictor $f$ by optimizing a Granger-causal objective. CXPlain can also estimate the uncertainty of features importance by calculating confidence interval (CI).

\subsubsection{Counterfactual Explanation}

Counterfactual explanation is the example-based model-agnostic method which generates new instances that would change the models' prediction. The prominent example \cite{grath_interpretable_2018} in this research is that one person $x$ with the annual income $a$ and the current balance $b$ has been rejected a loan by the financial institution, so how she/he can change her/his income and balance to $a^{\prime}$ and $b^{\prime}$ in order to receive the loan. Given the set of points $P$, in order to generate the set of counterfactual samples $F$, the objective function of counterfactual explanation \cite{wachter_counterfactual_2018} is to optimize the following function:

\begin{equation}
\begin{split}
    &\arg\min_{x}\max_{\lambda} (\lambda\cdot(\hat{f}(x^\prime)-y^\prime)^2+d(x,x^\prime))\\
    & d(x_i, x^\prime) = \sum_{k \in F} \frac{\mid x_{k} - x_k^{\prime} \mid}{MAD_{k}}\\
    & MAD_{k} = \underset{(j \in \text{P})}{\text{median}}(|X_{j,k} - \underset{(l \in \text{P})}{\text{median}}(X_{l,k})|)
\end{split}
\label{eqn:distance_function}
\end{equation}
where $x$ is an original instance, $x'$ is the counterfactual instance which close to $x$, $y^{\prime}$ is the target class label for $x'$, $\lambda$ is the regularized parameter, $d(x, x^\prime)$ denotes the distance between the original instance and the counterfactual samples, $MAD_{k}$ is the median absolute deviation for feature $k$. 

Grath et al.~\cite{grath_interpretable_2018} extend $d(x, x^\prime)$ in Eq. \eqref{eqn:distance_function} by adding a weight vector $\Theta$. The vector $\Theta$ is used to evaluate models' feature importance, and can be obtained by many algorithms such as K-Nearest Neighbors or global feature evaluation. Dhurandhar et al. \cite{dhurandhar2018explanations} combine the loss function generated from Convolutional AutoEncoder, while Arnaud \cite{van_looveren_interpretable_2020} uses the prototypes function to ensure that the generated perturbation falls into the same distribution with the original data as well as increasing the computational speed without tuning too many parameters. Additionally, the counterfactual samples should be as diverse as possible; the study \cite{mothilal_explaining_2020} proposes to use determinant of kernel matrix to illustrate this property. 

To empower the capability of counterfactual explanations, constraints are considered in optimization problem of counterfactual explanation. Take for example, a person cannot decrease his age, or change his race and skin color. Recent work~\cite{ustun2019actionable,russell_efficient_2019} adopt Mixed Integer Programming (MIP) formulation to deal with categorical, numeric and mixed data type. Meanwhile, Artelt et al. \cite{artelt2020convex} propose convex density constraints to generate counterfactual located in a region of the data space. Specifically, the density constraint $\hat{p}_{y} \ge \delta$ denoted by a kernel density estimator or a Gaussian mixture model is added into the distance function $d(x,x')$.

CERTIFAI \cite{sharma2019certifai} proposed by Sharma et al. as a novel and flexible approach which can be used in any type of data. CERTIFAI uses the customized genetic algorithm to choose individuals that have the best fitness scores defined as follows.

\begin{equation}
\label{eqn:fitness}
\begin{split}
&fitness = \frac{1}{d(x,x^{\prime})}\\
d(x,x^{\prime}) &=\begin{cases}
\frac{n_{x^{\prime} o n}}{n} l_1(\mathbf{x}, \mathbf{x^{\prime}})+\frac{n_{c a t}}{n} simp(\mathbf{x}, \mathbf{x^{\prime}}) & \text{tabular data}\\
\frac{1}{SSIM(x,x^{\prime})} & \text{image data} 
\end{cases}
\end{split}
\end{equation}
For tabular data, CERTIFAI chooses $l_1$ norm for continuous features and a simple matching distance for categorical features (simp). For image data, Structural Similarity Index Measure (SSIM) \cite{1284395} measures the similarity of what humans consider. $n_{con}$ and $n_{cat}$ are the number of continuous features and categorical features, respectively.

Instead of identifying the minimum changes leading to the desired outcome, a new line of counterfactual explanations provides feasible paths to transform a selected instance into one that meets a certain goal.
FACE \cite{poyiadzi2020face} proposed by Poyiadzi et al. constructs a graph over the data points with the weights illustrating the feasible degree to transit between two vertices. FACE thereafter can be solved by the \emph{Disjstra} algorithm to find the shortest path from the original instance to the counterfactual one.

\subsection{Visualization of Causal Effect}

Visualization-based method is another commonplace approach for quick understanding what the models have learned. Partial dependence plot (PDP) \cite{goldstein2015peeking} depicts the marginal effect of features into the predicted outcomes. The partial dependence function is defined as:

\begin{equation}
    \hat{f}_{x_S}(x_S)=E_{x_C}\left[\hat{f}(x_S,x_C)\right]=\int\hat{f}(x_S,x_C)p(x_C)dx_C
\end{equation}

Zhao et al.\cite{doi:10.1080/07350015.2019.1624293} use Partial dependence plot (PDP) an its extension called Individual Conditional Expectation (ICE) to extract the causal information from machine learning model. These visualization tools allow to measure the predictions' change after making an intervention, which can help to discover the features' causal relationship.

\section{Evaluation}
Evaluation in causal interpretability is an extremely difficult task, at least in the current stage, since there are nearly no grouthtruth data to evaluate the methods' performance. Evaluation for traditional interpretable machine learning evaluation can be classified into three categories~\cite{doshi2017towards}: application-based, human-based and function-based. We apply the same category and focus on evaluations that can be used in causal interpretability. 

\subsection{Application-based}

In real-world scenario where the machine learning model is deployed to assist experts, application-based evaluation illustrates how well the models provide explanations to human experts for improving their performance in specific tasks. Take for example, a randomized experiment \cite{10.1145/2876034.2876042} is conducted among a group of learner to solve the problems. They then rate the explanation generated by the machine learning models. With the assistance of models, the performance of people in different tasks is proved to be improved.

\subsection{Human-based}


Human-based evaluation methods refer to evaluate the performance of interpretable models with the assistance of human. 
Madumal et al. \cite{madumal2019explainable} generate explanation for the reinforcement learning. They implement an RL agent, and conduct an experiment running on StarCraft II, a strategic game, with 120 participants. \emph{Explanation Satisfactory Scale} \cite{hoffman2018metrics} is defined as the degree of human understanding of the AI system to measure the quality of generated explanations. 

\subsection{Function-based}
Functional-based evaluation methods can be carried out without the assistance of human to evaluate the performance of the explanation model. There are some evaluation procedures for different techniques in Section IV:
\subsubsection{Causal Interpretability for DNN}
The lack of ground truth for feature effect makes it challenging to evaluate the performance of causal effect estimation. Chattopadhyay et al. \cite{chattopadhyay2019neural} compare the salient map \cite{10.1023/A:1012460413855} generated by causal attribution method with Integrated Gradient \cite{sundararajan2017axiomatic}. Harradon etc al. \cite{harradon2018causal} identify the components having the significant causal effects into the individual prediction. Specifically, they conduct the experiments in three different architectures VGG 19 in Birds200, VGG 16 and 6-layer cov network applied in Inria dataset. Thereafter, they make a query for an individual input, and then visualize top $k$ variables according to their causal effect.

\subsubsection{Counterfactual Explanations}
A previous research \cite{mothilal_explaining_2020} suggests that there are three main metrics to evaluate the counterfactual explanation: \emph{proximity}, \emph{diversity} and \emph{sparsity}. The \emph{proximity} is to reflect the similarity between the CF examples and the original one which was calculated as the mean proximity all over the examples. Meanwhile, the \emph{diversity} measures the mean of the distances between the pairs of samples, ensuring that the generated instances should be as diverse as possible. Finally, the \emph{sparsity} is the average number of changes converting CF examples to the original one. 
\begin{equation}
\begin{split}
    \text{proximity} &= \frac{-1}{k} \sum_{i=1}^{k}dist(x_{cf_i},x)\\
\text{diversity} &= \frac{1}{C_{2}^{k}}\sum_{i=1}^{k-1} \sum_{j = i + 1}^{k}{dist(x_{cf_i}, x_{cf_j})}\\
\text{sparsity} &= 1 - \frac{1}{k\cdot d}\sum_{i=1}^{k} \sum_{l=1}^{d}1[x_{cf_i}^l \not\equiv  x_i^l]
\end{split}
\end{equation}
with $x_{cf}$ and $x$ are the counterfactual samples and original instance, respectively, $dist(x_{cf_i}, x_{cf_j})$ illustrates the distance between two generated counterfactual instances, $d$ is the number of input features, $k$ is the number of counterfactual samples to be generated.

\begin{table*}[!htb]
\centering
\caption{Toolbox for causal analysis}
\label{fig:toolbox}
\resizebox{\textwidth}{!}{\begin{tabular}[t]{lccc}
\toprule
Library & Feature & Algorithms & License \\
\midrule

DoWhy \cite{dowhy}   & \makecell{Individual treatment effect estimation} & \makecell{Prospensity score matching \cite{dehejia2002propensity} \\ Stratification \cite{frangakis2002principal}} & Microsoft  \\ 

EconML \cite{econml} & \makecell{Individual treatment effect estimation \\ Interpreter of the causal model} & \makecell{Double Machine Learning \cite{chernozhukov2016double} \\ Orthogonal Random Forests \cite{oprescu2018orthogonal, foster2019orthogonal} \\ Meta-Learners \cite{RN165} \\ Deep Instrumental Variables} & Microsoft  \\ 

Causal ML \cite{chen2020causalml} & Individual treatment effect estimation & \makecell{Meta-learners \\ Uplift modeling \cite{2017arXiv170508492Z, radcliffe2011real}}  & Uber  \\ 

Causal discovery toolbox \cite{2019arXiv190302278K} & Causal relationship discovery & \makecell{Graph Inference \\ Pairwise inference}  & ElementAI \\ 

CausalNex  & \makecell{
    Learn causal structures \\ Estimate the effects of potential interventions using data.
}  & Using Bayesian Networks for Causal Inference & QuantumBlack Labs  \\


TIGRAMITE &  Causal discovery for time series datasets  & PCMCI \cite{Rungeeaau4996}, Generally \cite{doi:10.1063/1.5025050}, CMIknn \cite{runge2017conditional}, Mediation class \cite{runge2015identifying,runge2015quantifying}  & GNU General Public \\ \bottomrule

\end{tabular}}
\end{table*}

\section{Open questions and discussions}

The need of explaining and interpreting models becomes highly critical along with the growing popularity of deep learning and automated machine learning. Although, there are currently several studies in this field, several open problems still remain unresolved.

\emph{1) Counterfactual explanation in classification tasks.} There are a plethora of constraints, especially features' causal relationship, should be taken into consideration when adopting counterfactual explanation. Take for example, the counterfactual explanation cannot recommend the users to change sensitive and discriminative features such as race and gender in order to be accepted by the system. Therefore, its reasonability and feasibility should be discovered and investigated more strictly. 

\emph{2) Counterfactual explanation in recommendation system and time series data.} Although recommendation system gains the immense popularity these days, there are not many studies working on counterfactual explanation for such system. How we can make an intervention into human actions to enable the system to change their recommended items still remains an open question. Meanwhile, regarding time series data, it is also interesting to discover that what the model would change its prediction if we change something in the past.  

\emph{3) Causal reasoning in knowledge graph.} Knowledge graph is recently utilized as an effective tool in several tasks such as recommendation system, knowledge extraction, classification, etc. Instead of embedding the knowledge graph as the latent features, Xian et al. \cite{xian2019reinforcement} state that the true intelligent recommendation systems have to own the ability to recommend their items based on their causal reasoning.

\emph{4) Explanation understandable by non-experts.} A number of recent methods frequently provide the explanations to experts and researchers rather than the end-users. Therefore, another challenge is to generate explanation under the form such as rules, natural language, images, etc which can allow nonprofessional people to catch up with machine learning model behaviors.

\section{Conclusion}
Interpretable machine learning is expected to become a mainstream topic in the foreseeable future. This paper provides the desiderata and brief overview of causal inference, followed by the causality based interpretable machine learning. We present two main causal approaches for interpretable machine learning including feature importance estimation, causal effects of model components, and counterfactual explanation. Finally, we has discussed several potentially unresolved problems in this field which open opportunities for researchers to work in.  

In machine learning, the more data the better. However, in causal inference, the more data alone is not yet enough. Having more data only helps to get more precise estimates, but it cannot make sure these estimates are correct and unbiased. Machine learning methods enhance the development of causal inference, meanwhile, causal inference also helps machine learning methods. The simple pursuit of predictive accuracy is insufficient for modern machine learning research, and correctness and interpretability are also the targets of machine learning methods. Causal inference is starting to help to improve machine learning, such as recommender systems or reinforcement learning.


\bibliographystyle{./IEEEtran}
\bibliography{interpretable_ref}


\end{document}